\documentclass{article}

\usepackage{arxiv}

\usepackage[utf8]{inputenc} 
\usepackage[T1]{fontenc}    
\usepackage{hyperref}       
\usepackage{url}            
\usepackage{booktabs}       
\usepackage{amsfonts}       
\usepackage{nicefrac}       
\usepackage{microtype}      
\usepackage{lipsum}
\usepackage{graphicx}
\graphicspath{ {./images/} }

\usepackage{microtype}

\usepackage{multirow}
\usepackage{amsmath}
\usepackage{comment}
\usepackage{graphics}
\usepackage{enumitem}
\usepackage{ulem}
\usepackage{subcaption}
\usepackage{natbib}
\setcitestyle{authoryear,round,citesep={;},aysep={,},yysep={;}}
\renewcommand{\cite}{\citep}

\font\titlefont = pplb at 16pt
\title{\titlefont On Initializing Transformers with Pre-trained Embeddings}

\author{
 Ha Young Kim \\
  Department of Computer Science\\
  State University of New York, Korea\\
  \texttt{hayoung.kim@stonybrook.edu} \\
   \And
 Niranjan Balasubramanian \\
  Department of Computer Science\\
  Stony Brook University\\
  \texttt{niranjan@cs.stonybrook.edu} \\
  \And
 Byungkon Kang \\
  Department of Computer Science\\
  State University of New York, Korea\\
  \texttt{byungkon.kang@sunykorea.ac.kr} \\
}

\begin{document}
\maketitle
\begin{abstract}
It has become common practice now to use random initialization schemes, rather than the pre-trained embeddings, when training transformer based models from scratch. Indeed, we find that pre-trained word embeddings from GloVe, and some sub-word embeddings extracted from language models such as T5 and mT5 fare much worse compared to random initialization. This is counter-intuitive given the well-known representational and transfer-learning advantages of pre-training. Interestingly, we also find that BERT and mBERT embeddings fare better than random initialization, showing the advantages of pre-trained representations. In this work, we posit two potential factors that contribute to these mixed results: the model sensitivity to parameter distribution and the embedding interactions with position encodings. We observe that pre-trained GloVe, T5, and mT5 embeddings have a wider distribution of values. As argued in the initialization studies, such large value initializations can lead to poor training because of saturated outputs. Further, the larger embedding values can, in effect, absorb the smaller position encoding values when added together, thus losing position information. Standardizing the pre-trained embeddings to a narrow range (e.g. as prescribed by Xavier) leads to substantial gains for Glove, T5, and mT5  embeddings. On the other hand, BERT pre-trained embeddings, while larger, are still relatively closer to Xavier initialization range which may allow it to effectively transfer the pre-trained knowledge.
\end{abstract}


\section{Introduction}

The transformer \cite{NIPS2017_3f5ee243} remains dominant as being the underlying architecture for state-of-the-art large language models \cite{LLMSurvey}. Because of the massive amount of resources and time required to train these models, any types of optimizations would be beneficial to save costs. Therefore, when training future transformer-based language models, initializing the right parameter values can be one of many methods that can help train the model more effectively. There are many good reasons for initializing the embedding layer with pre-trained word vectors such as Glove~\cite{pennington-etal-2014-glove}, Word2Vec~\cite{DBLP:journals/corr/abs-1301-3781}, and more recently sub-word vectors \cite{bojanowski-etal-2017-enriching} from language models such as BERT \cite{devlin-etal-2019-bert}. For one, they have been shown to capture a range of useful knowledge including lexical, syntactic, and even some types of factual relations~\cite{pennington-etal-2014-glove}. Also, from a transfer learning perspective, pre-trained parameters can help improve training effectiveness and convergence in downstream tasks, especially when training data is limited. Indeed, pre-trained token vectors\footnote{In this paper, we refer token embeddings to both sub-word and word embeddings} have been shown to be useful for a wide-range of applications in other non-transformer based models~\cite{kocmi-bojar-2017-exploration,kim-2014-convolutional,collobert2011natural,qi-etal-2018-pre,lample-etal-2016-neural}. However, when training transformer based models from scratch, it has become standard practice to prefer randomly initialized  embeddings over pre-trained embeddings. Our experiments also indicate that random initialization performs better than pre-trained word embeddings, while for sub-word embeddings the trends are not consistent. \textit{What makes certain pre-trained embeddings ineffective in transformers?}

To answer this question, we investigate two possible factors.
\begin{enumerate}[label=\textbf{\arabic*.}, leftmargin=*, noitemsep] 
    \item \textbf{Model sensitivity to parameter distribution}: Deep neural networks, including transformers, are sensitive to the variance of the parameters for proper gradient flow \cite{glorot2010understanding}. Studies from \citet{glorot2010understanding} have shown a suitable variance range for parameter initialization to enable better gradient flow and faster convergence during training. The random initialization scheme utilizing such a range is known as Xavier initialization and is standard practice in training deep neural networks. In effect, it restricts the model parameter values, including those in token embeddings, to be zero centered and within a narrow variance range~\footnote{\citet{he2015delving} also introduced a similar initialization scheme, which accounts for non-linear activation in Rectified Linear Units. In this work, we focus only on Xavier initialization for settings without ReLU activation.}. Pre-trained token vectors, however, are not necessarily subject to these distributional constraints and may perform poorly if the variance and mean does not lie within a preferred narrow range.  
    
\item \textbf{Interactions with Position Encodings}: Token embeddings in transformers are directly added to the positional encodings in order to track position information. If the variance of the word embeddings and positional encodings are widely different, then one type of information might dominate the other when added together. We refer to this phenomenon in the paper as "absorption", such as when the pre-trained embeddings with much higher variance "absorb" the positional encodings.  
\end{enumerate}
To test the impact of these factors, we conduct an empirical study comparing various pre-trained/random embedding initializations on four tasks --- machine translation (de-en for both Multi30k~\cite{elliott-etal-2016-multi30k} and IWSLT2017 \cite{cettolo-etal-2017-overview}, sentiment analysis (SST2) \cite{socher-etal-2013-recursive}, and natural language inference (MNLI) \cite{williams-etal-2018-broad}. We tested GloVe for pre-trained word embeddings, and for sub-word embeddings, they include those directly obtained from the embedding layers of BERT, multilingual BERT (mBERT) \cite{devlin-etal-2019-bert}), T5 \cite{raffel2020exploring}, and multilingual T5 (mT5) \cite{xue-etal-2021-mt5}. The experimental results support three key findings:

\textbf{Finding 1:} Pre-trained embeddings, such as GloVe, T5, and mT5, with variance many orders of magnitude higher than the variance of Xavier initialized embeddings tend to fare worse. Meanwhile, BERT and mBERT embeddings with a more similar range to Xavier performs on par or better than Xavier. Although the exact range for the optimal performance is not known, the Xavier initialization range have been widely used and tested to perform well. Therefore, pre-trained embeddings that do not fall within the Xavier specifications tend to be less effective. We know the embedding variance is important as experiments show that standardizing GloVe, T5, and mT5 embeddings to match the variance of the Xavier initialization scheme lead to substantial improvements. However, standardizing BERT and mBERT embeddings usually either made a neutral or negative effect to the performance.\\
\textbf{Finding 2:} Interactions of pre-trained embeddings with the position embedding have a two-way effect. (a) Adding pre-trained embeddings to position embeddings can reduce the impact of position embeddings, since the range of pre-trained embeddings is much larger. 
(b) Adding position embeddings greatly alters the word-word relations encoded by the Xavier standardized pre-trained vectors. Despite this, the residual structure returns small but consistent benefits. \\
\textbf{Finding 3:} Pre-trained embeddings do carry merit to the model performance in terms of semantic information. This is apparent in experiments where shuffling the elements of a pre-trained embedding layer degrades the transformer performance consistently even though the distribution remains untouched. In addition, standardizing certain pre-trained embeddings to match Xavier distribution sometimes yield small but consistent gains over Xavier initialized embeddings alone. 
\section{Related Work}
\label{section:related_work} 

While we could not find work that specifically studies the effects of inserting static pre-trained word embeddings into transformers, there are several studies showing the benefits of transferring pre-trained sub-word embeddings from language models with some modifications in between. A different form of embedding initialization in transformers has been mentioned in cross-lingual knowledge transfer for efficiently training resource-demanding language models~\citep{minixhofer-etal-2022-wechsel}. Instead of inserting pre-trained word embeddings learned from a separate algorithm like GloVe, the embeddings are transferred from another pre-trained transformer, therefore not needing to worry about discrepancies in embedding distribution. \citet{minixhofer-etal-2022-wechsel} devised a method to train sub-word embedding-based monolingual models more efficiently by initializing each sub-word embedding as a combination of the $k$-nearest sub-word embeddings in another pre-trained language model. \citet{dobler-de-melo-2023-focus} initialized the target model embeddings for tokens not found in the source model’s vocabulary by the weighted average of the most similar overlapping tokens embeddings, while the overlapping tokens’ embeddings are directly copied to the target model. While some of these works may use static embeddings like \citet{bojanowski-etal-2017-enriching} to select the embeddings from the source language model, the static embeddings themselves are not directly inserted into the model itself.

\section{Embedding Initializations}
\label{sec2.5:init}
We posit that distributional differences of pre-trained embeddings can be a key factor for their poor performance on transformers. Therefore, we will analyze the following embedding initializations in this paper. 

These methods vary in how the entries of the embedding layer $X=\{x_{ij}\}\in\mathbb{R}^{N\times D}$ of $N$ number of $D$-dimensional vectors are initialized. We denote random initializations by $X_x$ (for Xavier), and pre-trained embedding initializations by $X_p$. In practice, N is also the vocabulary size or number of tokens the model holds.

\subsection{Random Embeddings} \label{random_embeddings}

\textbf{Xavier~\cite{glorot2010understanding}}: Each $x_{ij}$ is sampled from the uniform distribution $U(-\sqrt{6/(N+D)}, \sqrt{6/(N+D)})$ in an i.i.d. manner. As mentioned in \cite{glorot2010understanding}, since the variance depends on the forward and backward pass gradients, the restricted range keeps the variance constant across layers to prevent exploding or vanishing gradients. The range is inversely proportional to $\sqrt{N}$ and in most cases, $N>D$, meaning that a larger vocabulary size equates to a smaller range. With such a dependency on the vocabulary size, there are cases when this initialization may not be optimal especially when only a small portion of the vocabulary is being used.

\subsection{Pre-trained Word Embeddings} \label{word_embeddings}
\textbf{GloVe ~\cite{pennington-etal-2014-glove}}: We use the GloVe word embedding vectors that are trained by the following objective.
\begin{equation*}
\label{eq:pre-trained}
 J(\mathbf{w},\mathbf{b}) = \sum_{i,j=1}^{V} f(x_{i,j})(w_i^{T}\Tilde{w}_j + b_i + \Tilde{b}_j - \log x_{i,j})^2
\end{equation*}
Here, $w_i$ and $\Tilde{w}_i$ are the word and context vectors, respectively, and $b$ and $\Tilde{b}$ are the respective bias parameters.
Note that unlike the Xavier initialization, the above objective does not constrain the range of the vectors and thus can, in principle, be unbounded. In practice, the range of the values depend upon the initialization of the word vectors during Glove training and the size and nature of the training data itself. For our translation experiments, we used GloVe embeddings pre-trained on German Wikipedia from Deepset\footnote{German embeddings downloaded from \url{https://www.deepset.ai/german-word-embeddings}} for the encoder, while the decoder is loaded with GloVe embeddings pre-trained on Common Crawl containing 840 billion tokens.\footnote{English embeddings downloaded from \url{https://nlp.stanford.edu/projects/glove/}} The remaining non-translation experiments used the same 840B token pre-trained embeddings.

\subsection{Pre-trained Sub-word Embeddings} \label{subword_embeddings}
To form the sub-word embedding layer, we select $N$ specific and non-overlapping vectors $y_i \in\mathbb{R}^{D}$ directly from a pre-trained language model's own embedding layer $Y_l \in\mathbb{R}^{V\times D}$ where $V>N$ and $X_p \subseteq Y_l$. The selected vectors are those mapped to sub-word tokens found in the vocabulary of the pre-trained model and our own transformer. The multilingual variants (denoted by the '$m$' prefix) of each pre-trained language models' embeddings are used in translation tasks only while the original models' embeddings are tested in remaining classification tasks. This separation is due to the original models being trained mostly using English corpora where semantic information in other languages are missing. \\
\textbf{BERT ~\cite{devlin-etal-2019-bert}}: BERT is an encoder-only transformers model, where inputs are processed similarly to the original transformers. The token embeddings are added with positional and sentence embeddings before being fed into the encoder and in order to preserve information from all three embeddings, their variances are thus bounded by each other. We use BERT embeddings for classification tasks only. \\
\textbf{mBERT ~\cite{devlin-etal-2019-bert}}: mBERT is a BERT model pre-trained on a corpora of 104 languages, therefore using the same architecture where the same bounding effect applies to its embeddings. \\
\textbf{T5 ~\cite{raffel2020exploring}}: The T5 model is nearly identical to the vanilla transformers, using an encoder-decoder architecture with a notable difference being that it uses relative position embeddings \cite{shaw-etal-2018-self}. Instead of adding the position information directly to the token embeddings, they are added as scalars inside the attention layer. The token embeddings are directly fed into the model and thus are not bounded by position information. \\
\textbf{mT5 ~\cite{xue-etal-2021-mt5}}: mT5 is a multilingual variant of T5 that is trained on an extended Common Crawl dataset covering 101 languages. Because of its identical model architecture to T5, its distribution range is similar to T5. 
\subsection{Modification of Pre-trained Embeddings from Section \ref{word_embeddings} and \ref{subword_embeddings}}
\label{modified_embeddings}

The following initialization methods essentially standardizes (shifts the mean and standard deviation) $X$ to match another embedding layer's distribution. By default, we refer "standardization" as shifting the mean and standard deviation of pre-trained embeddings to the ones Xavier possesses. \\
\textbf{Pre-trained\_Xavier (proposed)}: Here, we propose to transform $X_p$ via the following expression: 
\begin{equation}
\label{eq:xavier_standard}
x^*_{i,j} = (x_{i,j} - \overline{X_p}) \times \frac{\sigma_x}{\sigma_p}
\end{equation}
where $\overline{X_p}$ is the mean of the values in $X_p$ and $\sigma_x$ and $\sigma_p$ are the (sample) standard deviations of the Xavier- and Pre-trained initialized vector components, respectively. 
We let $X_{px}=\{x^*_{i,j}\}$ denote this set. The rationale is to match the variance of the pre-trained vector distribution to that of Xavier-initialized ones while keeping the original semantic information as much as possible. That way, we can observe the effect of the adjusted range on the performance. \\
\textbf{Pre-trained\_Shuffled}: Given any set of pre-trained token vectors $X_p$, the rows and columns of $X_p$ are randomly shuffled. This should destroy the word-to-word semantic relations while preserving its statistical distribution to observe the level of impact semantic information has to the performance. Because there were cases where pre-trained embeddings perform better, we wish to see whether this advantage comes from having a more optimal mean and variance or its semantic information. If the shuffled pre-trained embeddings perform better than Xavier embeddings, then we conclude the pre-trained distribution \textit{itself} to be more optimal than Xavier's. Also, if both Pre-trained and Pre-trained\_Shuffled show similar performance, this indicates the pre-trained semantic information to be unhelpful. 

Transformer models need position information to be modeled via position encodings. The standard solution has been to simply add them directly to the word embedding inputs. As such, the initialized vectors are added to the positional encodings $P_{ij}$ of the same size after being scaled by $\sqrt{D}$ to be given as input to the transformer.

In later sections, we refer to "Pre-trained" embedding types by their names to specify which pre-trained embeddings were used and transformed (e.g. if using BERT embeddings, then the naming will be "BERT", "BERT\_Xavier", "Xavier\_BERT instead of the general "Pre-trained", "Pre-trained\_Xavier", "Xavier\_Pre-trained).

\section{Experimental Setup}

We performed a translation task on a vanilla encoder-decoder transformer~\cite{NIPS2017_3f5ee243} using the Multi30k and IWSLT2017 data sets, and also tested two classification tasks, namely MNLI and SST2, on an encoder-only transformer. \\
For Multi30k and IWSLT2017, the model with the lowest validation loss and its corresponding epoch is chosen to calculate the test data BLEU score with its standard deviation. For sub-word translation tasks, only the multilingual variants (mBERT and mT5) are used, while the remaining sub-word tasks use BERT and T5.\\
For MNLI, the highest validation match and mismatch accuracy achieved during training, along with their standard deviation ($\sigma$), are reported. For SST2, the highest validation accuracy with its standard deviation and corresponding epoch is reported. The labels of the SST2 and MNLI test data are not publicly available and therefore only the validation accuracy is reported. 

Each of these datasets is tested with three types of pre-trained embeddings (GloVe, BERT/mBERT, and T5/mT5). We denote a single task and its following embedding initializations tested as an \textit{experiment}. For experiments using GloVe embeddings, three different embedding initializations were tested, including Xavier from Section \ref{random_embeddings}, GloVe from Section \ref{word_embeddings}, and GloVe\_Xavier from Section \ref{modified_embeddings}. For the remaining sub-word experiments, the relevant embeddings from Section \ref{random_embeddings}, \ref{subword_embeddings}, and everything from Section \ref{modified_embeddings} were tested. Each embedding initialization was tested multiple times with random seeds and the results are averaged over repeated runs. Results for all 12 experiments are shown in Table \ref{table:multitasks} and further experiment details are available in Appendix~\ref{app:exp_details}. 

\begin{figure}[htbp]
\begin{center}
    \includegraphics[width=1.0\linewidth]{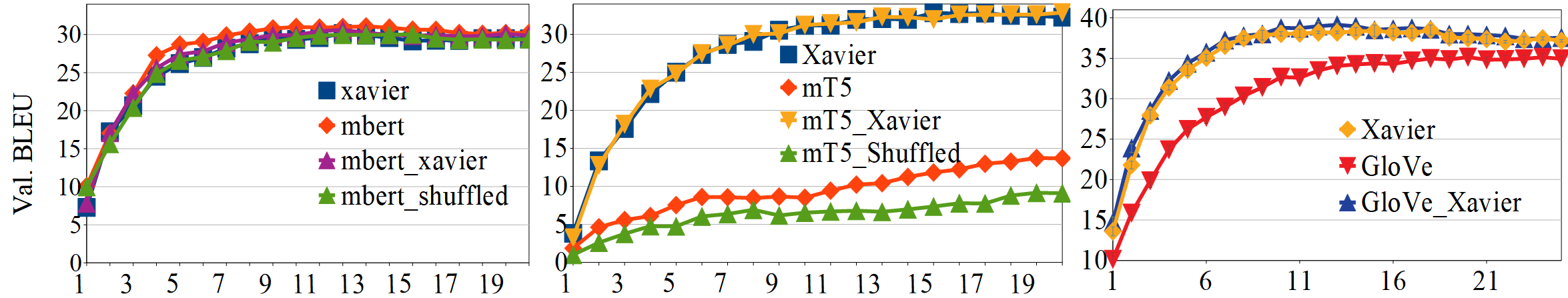}
    \caption{IWSLT2017 with mBERT (left), mT5 (center), and Multi30k GloVe (right) embeddings validation BLEU results throughout training epochs between various embedding initializations. Notice the variance of the pre-embeddings from Table \ref{table:sub_emb_statistics} ($\sigma_{mT5}>\sigma_{GloVe}>\sigma_{mBERT}$) and how that affects their relative performance gap to Xavier.}
    \label{fig:val_curve_subword}
\end{center}
\end{figure}

In addition, we repeated another Multi30k with GloVe embeddings experiment to plot the validation BLEU~\cite{papineni-etal-2002-bleu} curves (Figure \ref{fig:val_curve_subword} right). Five runs with random seeds were conducted for each embedding type and the average of the BLEU scores were recorded for each epoch. For the remaining charts in Figure \ref{fig:val_curve_subword}, we also plotted the IWSLT2017 validation BLEU scores, which are based off of the mBERT and mT5 experiments in Table \ref{table:multitasks}. 

All experiments had their vocabulary built from the training data. When building the vocabulary, the entire training data is tokenized and the resulting tokens appearing more than a set minimum frequency (5 for IWSLT2017 and 2 for everything else) are added to the vocabulary. All experiments using GloVe embeddings had the training dataset tokenized with the English and German spaCy\footnote{more information available from \url{https://spacy.io/usage/models}} \cite{spacy2} models. The remaining sub-word embedding experiments had its data tokenized with its respective language models. Because the pre-trained tokenizer and vocabulary sizes were different for each experiment, the Xavier initialization ranges vary and thus need to be evaluated separately for each experiment.

\section{Embedding Distribution Effects}
\label{exp:emb_dist}
\begin{table*}[th]
\begin{center}
\resizebox{\linewidth}{!}{%
\begin{tabular}{lll|ll|ll|ll}
\toprule
&\multicolumn{2}{c|}{\textbf{SST2}} & \multicolumn{2}{|c|}{\textbf{MNLI}} & \multicolumn{2}{|c|}{\textbf{Multi30k}} & \multicolumn{2}{|c}{\textbf{IWSLT2017}}\\
\hline
&Acc ($\sigma$)&Epoch&Match ($\sigma$)\slash Mismatch ($\sigma$)& Epoch& BLEU ($\sigma$) &Epoch & BLEU ($\sigma$) & Epoch\\ 
\hline
Xavier & 82.7 (0.5) &  1.3 & 60.4 (1.4)\slash 60.2 (1.8) & \textbf{3.5}\slash \textbf{3} & 37.3 (0.4) & \textbf{11.7} & 31.8 (0.4) & \textbf{12.8} \\
GloVe & 82.3 (0.7) &  3.8 & 57.7 (1.5)\slash 58.2 (1.2) & 6.5\slash7.5 & 34.1 (0.7) & 20.0 & 28.7 (0.2) & 20.0 \\
GloVe\_Xavier & \textbf{83.6} (0.2) &  \textbf{1} & \textbf{60.9} (0.0)\slash \textbf{61.1} (0.1) & 4\slash\textbf{3} & \textbf{38.7} (0.3) &  12.3 & \textbf{31.9} (0.5) & \textbf{12.8} \\
\hline
Xavier & 81.5 (1.0) &  1.4 & 60.9 (0.3)\slash 60.6 (0.7) & 3.7\slash 3.7 & 39.9 (0.6) & \textbf{7.8} & 31.4 (0.5) & 9.3 \\
BERT & \textbf{83.3} (0.5) &  1.6 & \textbf{66.3} (0.6)\slash \textbf{66.8} (0.6) & 4.3\slash5.0 & 39.4 (0.4) & 8.3 & \textbf{32.7} (0.3) & \textbf{9.0} \\
BERT\_Xavier & 81.6 (0.7) &  \textbf{1.0} & 60.3 (3.3)\slash 60.5 (0.6) & \textbf{3.3}\slash\textbf{3.0} & \textbf{40.2} (0.5) &  8.3 & 31.7 (0.3) & \textbf{9.0} \\
BERT\_Shuffled & 81.8 (0.6) &  1.6 & 61.0 (1.2)\slash 61.7 (1.2) & 6.3\slash 5.3 & 38.3 (0.6) &  9.0 & 32.1 (0.2) & 10.5 \\
\hline
Xavier & 81.4 (0.5) &  1.4 & 60.7 (0.1)\slash \textbf{61.3} (0.4) & \textbf{3.7}\slash \textbf{4.7} & 44.0 (0.3) & \textbf{8.3} & 32.4 (0.9) & \textbf{10.0} \\
T5 & \textbf{82.0} (0.6) &  3.6 & 53.6 (1.0)\slash 54.1 (1.0) & 10.0\slash9.7 & 35.9 (0.3) & 13.0 & 15.7 (0.2) & 19.3 \\
T5\_Xavier & 81.3 (0.3) &  \textbf{1.2} & \textbf{60.8} (0.5)\slash 60.8 (0.4) & 5.3\slash\textbf{4.7} & \textbf{44.9} (0.6) &  \textbf{8.3} & \textbf{32.9} (0.7) & 10.5 \\
T5\_Shuffled & 77.6 (1.0) &  4.2 & 50.1 (0.9)\slash 50.1 (0.9) & 9.3\slash 9.0 & 33.9 (0.9) &  13.3 & 11.5 (0.6) & 18.9 \\
\bottomrule
\end{tabular}
}
\end{center}
\caption{Results showing transformer performance difference across different tasks and embedding initializations, where each box represents an experiment. Values in bold indicate the best result for each experiment. Note that for BERT and T5 translation tasks, only the multilingual variants (mBERT and mT5) are used, although they are labelled as BERT and T5 in the table. All scores above are averaged and shown with their standard deviation $\sigma$.}
\label{table:multitasks}
\end{table*}

In this section, we discuss experimental results that support our three findings - namely, the embeddings' distributional effects (Finding 1), interaction with position encoding (Finding 2), and the semantic information of pre-trained embeddings (Finding 3).
\subsection{Word Embeddings} \label{section:emb_dist_word}

Supporting Finding 1, we observed pre-trained GloVe embeddings to have a much higher range and standard deviation than Xavier embeddings (Table~\ref{table:emb_statistics}), causing it to perform worse than Xavier-initialized embeddings across all tasks (Table \ref{table:multitasks}). We assume standard deviation to be the reason as standardizing it improved its performance by a few points in all tasks. Supporting Finding 3, GloVe\_Xavier embeddings had small, but noticeable performance gains compared to Xavier embeddings across all four tasks. This may be attributed to the residual similarity information in the GloVe\_Xavier embeddings (11.1\% Accuracy in Table \ref{table:2}). However, in an ideal scenario where position information is included while keeping similarity information, we hypothesize that it is possible to make more significant performance gains.

Figure \ref{fig:val_curve_subword} (right) shows a slightly different result than Table \ref{table:multitasks} where Pre-trained\_Xavier reached peak BLEU score slightly faster (39.1 at epoch 13) than Xavier (38.5 at Epoch 18). GloVe\_Xavier embeddings marginally outperform Xavier embeddings for each epoch, even though both have identical mean and standard deviation. With the same statistical parameters and their models running with the same hyperparameters, the difference in performance may be explained by the difference in their structures. 

\subsection{Sub-word Embeddings}

We observe a clear relationship between the variance of the sub-word embeddings (Table \ref{table:sub_emb_statistics}) and its performance across all tasks, which also supports Finding 1.
When standardizing T5 and mT5 embeddings to become T5\_Xavier, overall results from Table \ref{table:multitasks} indicate a significant performance increase in the MNLI, Multi30k, and IWSLT2017 tasks. This can be assumed by the much larger variance in T5 and mT5 embeddings that cause it to perform poorly ($11.52<\sigma_{T5}<22.99$ vs $0.01<\sigma_{Xavier}<0.02$ across all tasks) and how standardizing it greatly increases its performance. However, it is not exactly clear why T5 embeddings performed better than T5\_Xavier in SST2, despite its excessively wide distribution.  

On the other hand, BERT and mBERT embeddings consistently outperform the remaining embedding initializations in all experiments except for Multi30k, with MNLI showing the largest difference. This can be partially explained by BERT and mBERT embeddings having a similar variance to Xavier as shown in Table \ref{table:sub_emb_statistics} ($0.04<\sigma_{BERT}<0.05$ vs $0.01<\sigma_{Xavier}<0.02$ across all tasks). In addition, the increase in performance compared to random embeddings in general can also be explained by its semantic information. This is apparent when results show shuffled BERT embeddings to always perform worse than its original counterpart. The usefulness of this semantic information also varies for each task as shown by the variation in score differences between BERT and BERT\_Shuffled across tasks. Regardless, the consistent performance decrease when shuffling any pre-trained embeddings show that the semantic information is being utilized. The same pattern follows when shuffling T5 and mT5 embeddings, supporting Finding 3. 

Comparing the validation BLEU curves between mBERT and mT5 embeddings in Figure \ref{fig:val_curve_subword} also highlights the effect distribution has to the performance. With the much larger variance in mT5 embeddings, the validation BLEU for mt5 and mT5\_Shuffled gradually increases but overall struggles to improve as quickly as Xavier and mT5\_Xavier embeddings. Meanwhile, all embedding types for the mBERT experiment showed a similar performance, with mBERT consistently having higher Validation BLEU than others across all epochs. Both charts show a decrease in BLEU across all epochs when shuffling the embeddings, showing the advantage in semantic information. Sometimes, BERT's embedding distribution alone (without semantic information) can also be more favorable than Xavier's distribution. This can be seen when standardizing Xavier embeddings to match BERT's mean and variance marginally increased its performance in SST2 and IWSLT2017.

\begin{table}
\begin{center}
\resizebox{\linewidth}{!}{%
\begin{tabular}{lll|ll|ll|ll}
\hline
& \multicolumn{2}{c|}{\textbf{SST2}} & \multicolumn{2}{|c|}{\textbf{MNLI}} & \multicolumn{2}{|c|}{\textbf{Multi30k}} & \multicolumn{2}{|c}{\textbf{IWSLT2017}} \\ 
\hline
 & \textbf{Range} & \textbf{Mean ($\sigma$)} & \textbf{Range} & \textbf{Mean ($\sigma$)} & \textbf{Range} & \textbf{Mean ($\sigma$)} & \textbf{Range} & \textbf{Mean ($\sigma$)} \\
\hline
Xavier & -0.02,0.02 & 0 (0.01) & -0.01,0.01 & 0 (0.01) & -0.03,0.03 & 0 (0.02) & -0.01,0.01 & 0 (0.01) \\
GloVe & -3.9,3.9 & -0.01 (0.37) & -4.2,4.2 & 0 (0.37) & -3.9,4.8 & 0.05 (0.37) & -3.9,4.8 & 0 (0.33) \\
GloVe\_Xavier &  -0.13,0.13 & 0 (0.01) & -0.06,0.05 & 0 (0.01) & -0.12,0.26 & 0 (0.02) & -0.10,0.13 & 0 (0.01) \\ 
\hline
Xavier & -0.02,0.02 & 0 (0.01) & -0.02,0.02 & 0 (0.01) & -0.03,0.03 & 0 (0.02) & -0.02,0.02 & 0 (0.01)          \\
BERT & -0.95,0.73 & -0.03 (0.04) & -0.4,0.34 & -0.03 (0.04) & -0.61,0.22 & -0.01 (0.05) & -0.61,0.23 & -0.01 (0.05)             \\
BERT\_Xavier & -0.28,0.23 & 0 (0.01) & -0.08,0.08 & 0 (0.01) & -0.22,0.08 & 0 (0.02) & -0.15,0.06 & 0 (0.01)    \\
\hline
Xavier & -0.03,0.03 & 0 (0.01) & -0.02,0.02 & 0 (0.01) & -0.03,0.03 & 0 (0.02) & -0.02,0.02 & 0 (0.01)          \\
T5 & -792,332 & 0.12 (22.6) & -792,332 & 0.12 (23.0) & -113.5,94.5 & 0.11 (11.5) & -113.5,94.5 & 0.12 (11.7)           \\
T5\_Xavier & -0.51,0.22 & 0 (0.01) & -0.32,0.13 & 0 (0.01) & -0.16,0.14 & 0 (0.02) & -0.10,0.08 & 0 (0.01)   \\
\hline
\end{tabular}
}
\end{center}
\caption{
Statistics for the encoder embedding layers from Table \ref{table:multitasks}. \textbf{Range} is the minimum and maximum values of all elements in the embedding layer. \textbf{Mean} and \textbf{$\sigma$} are the sample mean and standard deviation of the embedding layer, respectively. Pre-trained\_Shuffled had identical distribution to Pre-trained. Note that for BERT and T5 translation tasks, only the multilingual variants (mBERT and mT5) are used, although they are labelled as BERT and T5 in the table. 
}
\label{table:sub_emb_statistics}
\end{table}

\section{Positional Encoding Effects}
\label{exp:pos_enc}
This section presents the experimental results that verify Finding 2; i.e., the interaction between embeddings and positional encoding. We discuss this effect on the translation task and the word analogy task.

\textbf{Translation.} To evaluate the effects of positional encoding, we repeated the GloVe Multi30k translation experiment from Section \ref{section:emb_dist_word} with the following embedding configurations: 1. whether embeddings are trainable 2. whether to disable positional encoding (directly feed input embeddings to the attention layer), and 3. the type of embedding initialized. All possible combinations of embedding configurations are tested and results are shown in Table~\ref{table:1}. The model with the lowest validation loss is then used to evaluate the BLEU score on the first thousand samples from the Multi30k test data. More experiment details are mentioned in Section \ref{app:exp_details} and hyperparameter settings can be found in Table \ref{table:hyper}. 

\begin{table}[htbp]
\begin{center}
\begin{tabular}{llll}
\hline
 & \textbf{Train\slash Pos.} & \textbf{Val. Loss (epoch)} & \textbf{BLEU ($\sigma$)}\\
\hline
\multirow{4}{*}{Xavier} 
&no\slash no& 1.98 (18.67) &	28.3 (0.08) \\
&yes\slash no&	1.93 (9.33)&	29.6 (0.53) \\
&no\slash yes&	1.77 (20.00)&	34.6 (0.5) \\
&yes\slash yes&	    1.61 (11.67)&	37.3 (0.42) \\
\hline
\multirow{4}{*}{GloVe} 
&no\slash no&	1.73 (18.00)&	31.2 (0.18) \\
&yes\slash no&	1.72 (19.00)&	31.9 (0.34) \\
&no\slash yes&	1.64 (19.67)&	33.9 (0.43) \\
&yes\slash yes&	    1.62 (20.00)&	34.1 (0.66) \\
\hline
\multirow{4}{*}{GloVe\_Xavier} 
&no\slash no&	1.73 (18.67)&	31.2 (0.19) \\
&yes\slash no&	1.85 (\textbf{9.00})&	31.2 (0.92) \\
&no\slash yes&	\textbf{1.53} (20.00)&	36.4 (0.23) \\
&yes\slash yes&	    1.55 (12.33)&	\textbf{38.7} (0.27) \\
\hline
\end{tabular}
\end{center}
\caption{\label{table1-caption}
Results showing difference in translation performance when enabling or disabling training and positional encoding. \textbf{Val. Loss} refers to the lowest validation loss achieved during training and its corresponding \textbf{epoch}. \textbf{Train} means whether the embedding layer can update its parameters during training and \textbf{Pos.} means whether positional encoding is enabled. \textbf{BLEU} is the score evaluated for the test data and $\sigma$ is its standard deviation. 
}
\label{table:1}
\end{table}

In Table \ref{table:1}, GloVe and GloVe\_Xavier embeddings performed similarly when turning off training and positional encoding, while Xavier embeddings didn't perform as well, indicating the usefulness of the pre-trained structure. For all types, only enabling position had a much higher performance gain compared to only enabling training, showing the importance of position information. However, we surmise position encoding isn't being learned as well in GloVe embeddings and this is evident when enabling positional encoding doesn't boost performance as much as the other two types. Finally, it struggles to converge for all configurations, showing that having a low standard deviation is important for convergence.

\textbf{Word Analogy Task.} The Word analogy task can be used to evaluate the number of correct word-to-word relations in a set of embedding vectors \cite{DBLP:journals/corr/abs-1301-3781}. To verify if position information is being learned and\slash or semantic similarity structure is being destroyed when adding positional encodings, we ran the word analogy task on various transformations of the GloVe embeddings. Table \ref{table:2} shows how semantic similarity information between embeddings are affected by scaling, standardization, and positional encoding addition.

From Table \ref{table:2}, adding positional encoding has no effect on the GloVe Embedding accuracy, because of the much higher standard deviation. Note that in the transformer input process, embeddings are scaled before adding positional encodings, so that the standard deviation differences are even more exaggerated (6.39 vs. 0.67 in Table \ref{table:emb_statistics}). However, once it's scaled down to Xavier's range and added with positional encodings (GloVe \& Xavier \& Pos. Enc.), the accuracy drops significantly. Accuracy also drops when adding GloVe to the scaled up position encodings (GloVe \& (Pos. Enc. $\times$ $\sigma_p/\sigma_x$)). This means position information is being encoded at the expense of destroying the structure. 

\begin{table}[htbp]
\begin{center}
\begin{tabular}{l|l}
\toprule
\textbf{Embedding} & \textbf{Total acc. (\%)}\\
\hline
GloVe & \multirow{5}{*}{1424\slash1631 (87.3)}  \\
GloVe \& Scaled  \\
GloVe \& Pos. Enc.  \\
GloVe \& Xavier-Std. \\
GloVe \& Xavier-Std \& Scaled \\
\hline
GloVe \& Xavier-std \& Pos. Enc. & 210/1631 (12.9) \\
GloVe \& (Pos. Enc. $\times$ $\sigma_p/\sigma_x$) & 181/1631 (11.1) \\
\bottomrule
\end{tabular}
\end{center}
\caption{\label{table2-caption}
Word-analogy task done on various types of modified target (decoder) GloVe embeddings. \textbf{Total acc.} is the number correctly predicted word-analogies divided by the total applicable questions. Applicable means that only the analogies where all words exist in the embedding vocabulary are chosen. Scaled means the embeddings are multiplied by $\sqrt{D}$, Pos. Enc. means positional encoding is added to the embeddings, and Xavier-std. means to transform the embeddings to match Xavier's standard deviation. Modifications are done in the order (left to right) it is labelled above. $\sigma_p$ and $\sigma_x$ refer to Equation \ref{eq:xavier_standard}.    
}
\label{table:2}
\end{table}  

Whether the worse performance in GloVe Embeddings is mainly due to higher standard deviation or position information not being learned is not certain. However, we can be sure that both position information and standard deviation are important to maintain performance and compromising either one hurts it.

\section{Limitations}
Our limitations include the lack of theoretical backing to explain the effects of adding positional encoding and the experiments only being tested on vanilla transformers. In addition, while embedding initialization affects a sizable portion of the total parameters in transformers depending on factors such as vocabulary size, there are limits to performance gains when the remaining non-embedding parameters are not being considered. 

With only a marginal performance increase using Pre-trained\_Xavier embeddings compared to Xavier initialization, its potential value needs further investigation. The storage space and computational costs required for pre-trained word vectors with lower-than-expected returns also calls for additional development.
Meanwhile, there are other many well-researched and developed forms of knowledge transfer with more effective training strategies. Depending on how much gains an uncompromised embedding including position and semantic information can provide, further studies are required to come to a conclusion on the value of initializing pre-trained embeddings.
\section{Conclusion}
\label{section:conclusion}
Pre-training has been key to the success of transformer-based models. \textit{A priori}, one would expect pre-trained word vectors to perform at least as well as, if not better than randomly initialized token embeddings, but that is not always so. This paper provides an explanation in terms of the distributional difference and the interactions with position encodings. We find that simply standardizing some pre-trained embeddings to the Xavier range can fix this discrepancy. 

For our future works, we can include running other tasks from the GLUE benchmark \cite{wang-etal-2018-glue}, conduct more studies on position encodings that do not alter the embeddings, such as relative positional encodings \cite{shaw-etal-2018-self} found in T5, and conduct more repeated tests to obtain reliable results. In addition, testing translation tasks on languages with flexible word-order can help us understand cases with less dependency on positional encoding. We can also experiment on other position-independent tasks to verify the usefulness of the pre-trained embedding structure when positional encoding is not added. The optimal case would be to find a method to retain both position and semantic similarity information without compromising either one.

\newpage 
\bibliography{references}

\appendix
\newpage
\section{Appendix}
\label{app:init}

\subsection{Transformer Input Process}
Following the procedure for feeding inputs using the baseline transformer model from \citet{NIPS2017_3f5ee243}, the word embeddings are first multiplied by a scalar value and then added by the corresponding positional encoding vector. Consider a sequence $X_i=\{x_1,...x_n\}$ of word vectors $x_i\in\mathbb{R}^{1\times D}$, where $D$ is the embedding dimension, that will be inputted into the transformer. Each vector $x_i$ is first scaled by a constant $\sqrt{D}$, and this scaled vector will be called $x_{i,scaled} = x_i \times \sqrt{D}$. The positional encoding is a lookup table $P\in\mathbb{R}^{L\times D}$, where $L$ is a user-defined maximum sequence length, that will be used to encode position information to the input sequence. The positional encoding value $P_{ij}$ in the $i$-th row and $j$-th column of $P$ is defined as follows: 

\begin{equation}
P_{ij} =
\begin{cases} 
      \sin\left(\frac{i}{10000^{(j/D)}}\right) & \text{if j is even} \\
      \cos\left(\frac{i}{10000^{(j/D)}}\right) & \text{if j is odd} \\
   \end{cases}
\end{equation}

Where $i$ is also interpreted as the position of the word in the sequence and $j$ refers to the $j$-th dimension in the word vector $x_i$. Therefore, $P_{ij}$ will be added to the to the $j$-th element in $x_i$, called $x_{ij,scaled}$ to get the final vector $x_{ij,encoded}$ that is scaled and encoded with position information:

\begin{equation}
x_{ij,encoded} = x_{ij,scaled} + P_{ij}
\end{equation}

This process is repeated for all elements in $x_i$ and all word vectors in $X_i$. After these processes, The sequence is then ready to be fed into the transformer. 

The scaling step is also included in  \citet{NIPS2017_3f5ee243}, but no details were given on the reason. However, because the embedding layer of the transformer was only intended for the much smaller range of Xavier-initialized embeddings, scaling the pre-trained embeddings causes its range to be substantially bigger than the [-1,1] range in the positional encodings. 


\subsection{Experiment Details}
\label{app:exp_details}

\textbf{Hyperparameters and overall settings.} \label{app:hyper} Table \ref{table:hyper} lists the hyperparameters for all tasks. All models in this experiment use the Adam optimizer, with $\beta_1 = 0.9$, $\beta_2 = 0.98$, and $\epsilon = 10^{-9}$, a dropout rate of 0.1, 8 heads (10 for GloVe), and 3 layers in either the encoder or decoder. The learning rates for all translation tasks are set to $2\times10^{-4}$, and the remaining tasks are set to $1\times10^{-4}$. Implementation is done in Pytorch \cite{NEURIPS2019_bdbca288} and data is obtained from Hugging Face Datasets \cite{lhoest-etal-2021-datasets}. The embedding dimension is set to $D=\{300,512,768\}$ for GloVe, T5, and BERT/mBERT/mT5 embeddings, respectively.

\textbf{Section \ref{exp:emb_dist} and \ref{exp:pos_enc}} \label{app:emb_dist} All non-embedding parameters are initialized with Xavier. For MNLI using GloVe embeddings, A hyperparameter grid search was performed for each embedding type, where each possible combination of hyperparameters and embedding type in the defined search space was run two times with random seeds. Only the best averaged results for a particular hyperparameter combination is shown in Table \ref{table:multitasks}. The number of random seed reruns for each embedding type were 3,4,5 for MNLI, Multi30k/IWSLT2017, and SST, respectively. The remaining experiments with pre-trained language models used sub-word embeddings copied from the model embedding layers using the process in Section \ref{subword_embeddings}

\begin{table}[htb]
\begin{center}
\resizebox{\linewidth}{!}{%
\begin{tabular}{l|lll|lll|lll|lll}
\hline
& \multicolumn{3}{c|}{\textbf{Multi30k}} & \multicolumn{3}{|c|}{\textbf{MNLI}} & \multicolumn{3}{|c|}{\textbf{SST2}} & \multicolumn{2}{|c}{\textbf{IWSLT2017}}\\
\hline
& GloVe & mBERT & mT5 & GloVe & BERT & T5 & GloVe & BERT & T5 & GloVe & mBERT & mT5  \\
\hline
Parameters & 11M & 39M & 39M & 26M & 29M & 17M & 6M & 18M & 9M & 25M & 62M & 67M \\
Epochs & 20 & 20 & 20 & 9 & 10 & 10 & 5 & 5 & 5 & 20 & 20 & 20  \\
Batch Size & 64 & 128 & 96 & 128 & 128 & 128 & 128 & 128 & 128 & 128 & 128 & 112 \\
Vocab size & 7.8k\slash5.9k & 6.3k\slash5.5k & 6.5k\slash5.5k &  79k & 26k & 23k & 13k & 11k & 8.8k & 27k\slash20k & 15k\slash16k & 18k\slash18k  \\
Train size & 29k & 29k & 29k & 393k & 393k & 393k & 67k & 67k & 67k & 206k & 206k & 206k \\
Min\slash Epoch & 1 & 0.7 & 0.7 & 3.5 & 11.2 & 7.2 & 0.5 & 1 & 0.8 & 3.3 & 7.8 & 9.5 \\
\hline
\end{tabular}
}
\end{center}
\caption{
Hyperparameters and training details for all of the tasks done in this experiment. Vocab size values refer to the encoder and decoder, respectively. Min\slash Epoch is the average time required to train for one epoch in minutes.}
\label{table:hyper}
\end{table}

\textbf{Word Analogy Task Details.} For Table \ref{table:2}, all embeddings are based off of the pretrained embeddings extracted from the target embedding layer. Only 1631 out of 19533 word analogy questions (8.35\%) were answered. The steps for reproducing Table \ref{table:2} results are: 1. build the vocabulary from the Multi30k dataset containing $N=5893$ most frequent words and initialize a 5893 $\times$ 300 sized embedding layer. 2. Insert the pre-trained embeddings corresponding to the 5893 words into the embedding layer 3. Perform the necessary modifications to the embedding layer (e.g. scaling or adding positional encodings\footnote{When adding positional encodings, for each row in the embedding layer, a row vector from the positional encoding matrix is randomly chosen and added to the embedding vector. The randomness is to evenly distribute the usage of all the positional encoding vectors and simulate an example encoded input sequence to feed to the transformer.} to them) and 4. Extract all the embeddings from the embedding layer and run the word-analogy task on them. 

\begin{table}
\begin{center}
\begin{tabular}{lll}
\hline
 & \textbf{Min, Max} & \textbf{Mean ($\sigma$)} \\
\hline
Xavier  & -0.027, 0.027 &	$\approx{0}$ (0.016)  \\
(Scaled) & -0.47, 0.47 &	$\approx{0}$ (0.27)  \\
\hline
GloVe  & -3.93, 4.82 &	0.053 (0.37)  \\
(Scaled) & -68.1, 83.4 &    0.91 (6.39)  \\
\hline
Pretrained\_Xavier & -0.17, 0.20 &	$\approx{0}$ (0.016)   \\
(Scaled) & -2.94, 3.51 &	$\approx{0}$ (0.27)   \\
\hline
Pos. Encoding & -1, 1 & 0.131 (0.67) \\

\end{tabular}
\end{center}
\caption{
Statistics for the encoder embedding layer with different initializations, where the embeddings are obtained from Section \ref{exp:pos_enc}. Min,Max are the minimum and maximum values of all elements in the embedding layer. (Scaled) means the embedding is scaled by $\sqrt{D}$, and this is shown to compare its range difference with the positional encoding since the scaled version is the actual one being added with the positional encoding. The range of Xavier initialization is obtained from the 5893 $\times$ 300 dimension embedding layer.
}
\label{table:emb_statistics}
\end{table}

\subsection{Other Embedding Initializations Tested}
Here, we show information and results for other embedding initializations tested, but not included in the main text due to its redundancy with other initializations or lack of explanations.

\textbf{Xavier\_Pre-trained} We transform $X_x$ to an expression similar to Equation \ref{eq:xavier_standard}:
\begin{equation}
\label{eq:pretrained_standard}
x^*_{i,j} = (x_{i,j} - \overline{X_x}) \times \frac{\sigma_p}{\sigma_x} + \overline{X_p}
\end{equation}
Where $\overline{X_x}$ is the mean of $X_x$, and the remaining symbols mean the same as those from Equation \ref{eq:xavier_standard}. We let $X_{xp}=\{x^*_{i,j}\}$ denote this set. This transformation allows $X_x$ to match its mean and standard deviation to $X_p$. Xavier initialization does not always have the optimal parameter distribution for larger embedding layers, so we wanted to check if the pre-trained distribution is more optimal than Xavier distribution by matching its mean and variance and seeing an increase in performance. Indeed, we found that standardizing Xavier embeddings to certain pre-trained ones such as BERT improve BLEU its performance slightly as shown in Table \ref{table:xavier_pretrained_perform}.

\begin{table}[htbp]
\begin{center}
\resizebox{\linewidth}{!}{%
\begin{tabular}{lll|ll|ll|ll}
\toprule
&\multicolumn{2}{c|}{\textbf{SST2}} & \multicolumn{2}{|c|}{\textbf{MNLI}} & \multicolumn{2}{|c|}{\textbf{Multi30k}} & \multicolumn{2}{|c}{\textbf{IWSLT2017}}\\
\hline
&Acc ($\sigma$)&Epoch&Match ($\sigma$)\slash Mismatch ($\sigma$)& Epoch& BLEU ($\sigma$) &Epoch & BLEU ($\sigma$) & Epoch \\
Xavier\_BERT & 82.0 (1.0) &  1.8 & 60.7 (0.3)\slash 60.5 (0.2) & 5.0\slash 4.3 & 37.9 (0.4) &  8.5 & 32.2 (0.5) & 10.8 \\
Xavier\_T5 & 79.2 (0.8) &  4.2 & 50.8 (0.1)\slash 51.0 (0.5) & 8.3\slash 8.7 & 34.3 (0.8) &  12.3 & 12.0 (0.7) & 19.5 \\
\bottomrule
\end{tabular}
}
\caption{Results showing trained transformer performances initialized with Xavier\_Pre-trained embeddings. Note that for BERT and T5 translation tasks, only the multilingual variants (mBERT and mT5) are used, although they are labelled as BERT and T5 in the table.}
\label{table:xavier_pretrained_perform}

\vspace{1cm}

\resizebox{\linewidth}{!}{%
\begin{tabular}{lll|ll|ll|ll}
\hline
& \multicolumn{2}{c|}{\textbf{SST2}} & \multicolumn{2}{|c|}{\textbf{MNLI}} & \multicolumn{2}{|c|}{\textbf{Multi30k}} & \multicolumn{2}{|c}{\textbf{IWSLT2017}} \\ 
\hline
 & \textbf{Range} & \textbf{Mean ($\sigma$)} & \textbf{Range} & \textbf{Mean ($\sigma$)} & \textbf{Range} & \textbf{Mean ($\sigma$)} & \textbf{Range} & \textbf{Mean ($\sigma$)} \\
\hline
Xavier\_BERT & -0.10,0.05 & -0.03 (0.04) & -0.10,0.05 & -0.03 (0.04) & -0.09,0.07 & -0.01 (0.05) & -0.09,0.07 & -0.01 (0.05) \\
Xavier\_T5 & -38.9,39.2 & 0.12 (22.6) & -39.7,39.9 & 0.12 (23.0) & -19.9,20.1 & 0.11 (11.5) & -20.1,20.3 & 0.12 (11.7)    \\
\hline
\end{tabular}
}
\end{center}
\caption{
Statistics for the encoder embedding layers from Table \ref{table:xavier_pretrained_perform}. \textbf{Range} is the minimum and maximum values of all elements in the embedding layer. Note that for BERT and T5 translation tasks, only the multilingual variants (mBERT and mT5) are used, although they are labelled as BERT and T5 in the table. 
}
\label{table:xavier_pretrained_emb_statistics}

\end{table}

\end{document}